\documentclass[10pt,twocolumn,letterpaper]{article}

\usepackage{cvpr}              %
\usepackage{booktabs}

\usepackage{xcolor}
\usepackage{amsmath}
\usepackage{amssymb}
\usepackage{graphicx}
\usepackage{xspace}
\usepackage{multirow}
\usepackage{comment}

\definecolor{turquoise}{cmyk}{0.65,0,0.1,0.3}
\definecolor{purple}{rgb}{0.65,0,0.65}
\definecolor{dark_green}{rgb}{0, 0.5, 0}
\definecolor{orange}{rgb}{0.9, 0.6, 0.1}
\definecolor{red}{rgb}{0.8, 0.2, 0.2}
\definecolor{darkred}{rgb}{0.6, 0.1, 0.05}
\definecolor{blueish}{rgb}{0.0, 0.3, .6}
\definecolor{light_gray}{rgb}{0.7, 0.7, .7}
\definecolor{pink}{rgb}{1, 0, 1}
\definecolor{greyblue}{rgb}{0.25, 0.25, 1}

\def \customparskip {.2em}
\renewcommand{\paragraph}[1]{\vspace{\customparskip}\noindent\textbf{#1}}

\newcommand{\loss}{\mathcal{L}}

\newcommand{\calN}{\mathcal{N}}

\newcommand{\expect}{\mathbb{E}}
\newcommand{\real}{\mathbb{R}}
\newcommand{\coord}{\mathbf{x}}
\newcommand{\error}{\mathrm{err}}
\newcommand{\support}{\mathcal{R}}
\newcommand{\importance}{Q}
\newcommand{\stopgrad}{\mathrm{sg}}

\newcommand{\params}{\boldsymbol{\psi}}

\newcommand{\noise}{\boldsymbol{\eta}}
\newcommand{\sampledist}{P}
\newcommand{\hardparam}{\alpha}
\newcommand{\gt}{f_\text{gt}}

\definecolor{cvprblue}{rgb}{0.21,0.49,0.74}
\usepackage[pagebackref,breaklinks,colorlinks,citecolor=cvprblue]{hyperref}
\usepackage{graphicx}
\usepackage{adjustbox}
\usepackage{multirow}
\usepackage{array}

\title{Accelerating Neural Field Training via Soft Mining}

\author{
  Shakiba Kheradmand\textsuperscript{1},
  Daniel Rebain\textsuperscript{1},
  Gopal Sharma\textsuperscript{1},
  Hossam Isack\textsuperscript{2}, \\
  Abhishek Kar\textsuperscript{2}, 
  Andrea Tagliasacchi\textsuperscript{3, 4, 5},
  Kwang Moo Yi\textsuperscript{1}
  \\
  \textsuperscript{1} University of British Columbia, 
  \textsuperscript{2} Google Research, \\
  \textsuperscript{3} Google DeepMind, 
  \textsuperscript{4} Simon Fraser University,
  \textsuperscript{5} University of Toronto
}

\begin{document}
\twocolumn[{%
\renewcommand\twocolumn[1][]{#1}%
\maketitle
\centering
\vspace{-1em}
\includegraphics[width=0.97\linewidth]{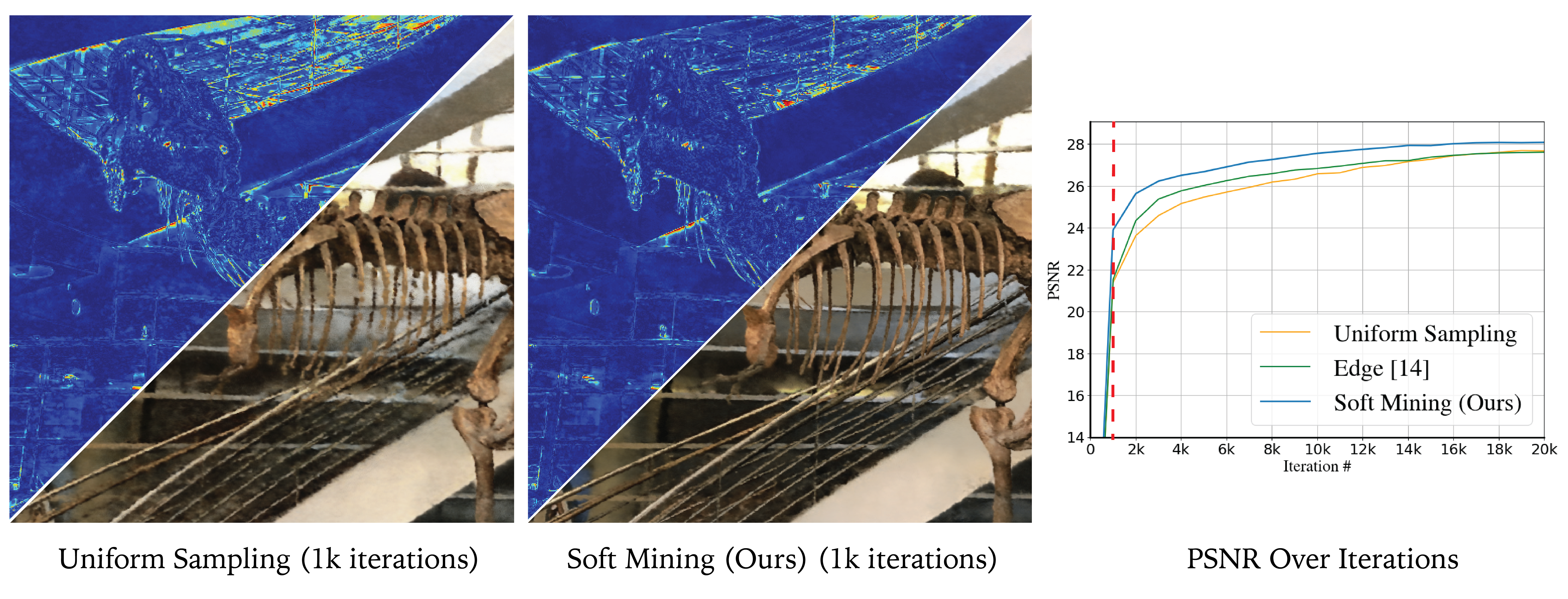}
\vspace{-1em}
\captionof{figure}{\textbf{Teaser:}
we introduce ``soft mining'' to accelerate neural field training. 
When applied to Neural Radiance Field (NeRF) training, our method significantly improves convergence.
We visualize the error maps (blue denotes low error and red denotes high error) and the rendered novel views for uniform sampling and our method.
We plot the convergence showing the Peak Signal-to-Noise Ratio (PSNR) for the corresponding scene. 
We render both images at 1k iterations of training, specified by the red dashed line in the {\bf (right)} graph.
Our method achieves the same PSNR significantly faster than the baselines.
\vspace{2em}
}
\label{fig:teaser}
}]
\vspace{-1mm}
\begin{abstract}
\vspace{-3mm}
We present an approach to accelerate Neural Field training by efficiently selecting sampling locations. 
While Neural Fields have recently become popular, it is often trained by uniformly sampling the training domain, or through handcrafted heuristics.
We show that improved convergence and final training quality can be achieved by a soft mining technique based on importance sampling: rather than either considering or ignoring a pixel completely, we weigh the corresponding loss by a scalar.
To implement our idea we use Langevin Monte-Carlo sampling.
We show that by doing so, regions with higher error are being selected more frequently, leading to more than 2x improvement in convergence speed. The code and related resources for this study are publicly available at \href{https://ubc-vision.github.io/nf-soft-mining/}{project page}.
\vspace{-4mm}
\end{abstract}    
\section{Introduction}
Neural fields~\cite{Xie22} have recently become popular due to their versatile nature, and their ease of integration with popular workloads like novel view synthesis~\cite{Mildenhall2020nerf}.
Neural fields map input coordinates to output values and are typically implemented as variants of multi-layer perceptrons~(MLP)~\cite{sitzmann2020siren, lindell2022bacon, Tancik2020fourierfeatures}.
They have demonstrated impressive capabilities in representing signals for 1D audio~\cite{sitzmann2020siren}, 2D images \cite{mueller2022instant, Tancik2020fourierfeatures}, 3D shapes \cite{Mildenhall2020nerf, Park_2019_CVPR}, and 4D light fields \cite{Suhail2022lfnr, yu2021plenoctrees}.
Beyond modeling and compressing signals, they have also been utilized to simulate physics~\cite{cai2021physics, pokkunuru2022improved}.

While neural fields provide impressive results, training them can be lengthy, e.g. on modern GPUs training the original Neural Radiance Field (NeRF)~\cite{Mildenhall2020nerf} for a single scene can take hours.
Researchers have since improved the training process through alterations to network architectures~\cite{mueller2022instant, tensorrf}, and through better loss functions \cite{barron2023zip}.

An angle that none of the papers above consider is how \textit{training batches} are formed.
That is, which pixels in the 2D image, or which rays in the radiance field are used for each optimization step.
These methods typically rely on simple \textit{uniform} sampling, which may lead to sub-optimal training performance.
For example, as shown for the NeRF training example in \Cref{fig:teaser},
if the signal has smooth regions such as walls, 
we can expect diminishing returns when sampling from those smooth areas as training progresses.
There have been some very recent attempts to address this problem either through a heuristic that focuses training to image edges~\cite{GAI2023104670}, or by analyzing images through quad-trees~\cite{zhang2022fast}.
However, these methods are hand-designed for NeRF and do not generalize to other neural field training.
Moreover, the gains that these methods provide are marginal, especially with real-world data, as we will show empirically.

In this work, we propose a principled method to improve the convergence of neural field training by improving the sampling mechanisms.
Akin to core research in image classification that introduced the concept of hard mining~\cite{shrivastava2016training, simo2015discriminative}, we propose to focus on `hard' samples within a neural field workload.
However, we empirically discovered that straightforward hard mining does not improve training.
We thus re-formulate neural field training as importance sampling and derive a relaxed `soft' hard mining formulation, analogous to the soft mining heuristic introduced for metric learning~\cite{wang2019deep}.
To implement our idea, inspired by the rich literature on Bayesian optimization for efficient sampling of distributions \cite{welling2011bayesian, blanchard2021bayesian, durmus2018efficient}, we opt to use Langevin Monte Carlo~(LMC) sampling~\cite{Brosse2018, mcmchandbook2012}.

\begin{figure}
    \centering
    \includegraphics[width=\linewidth]{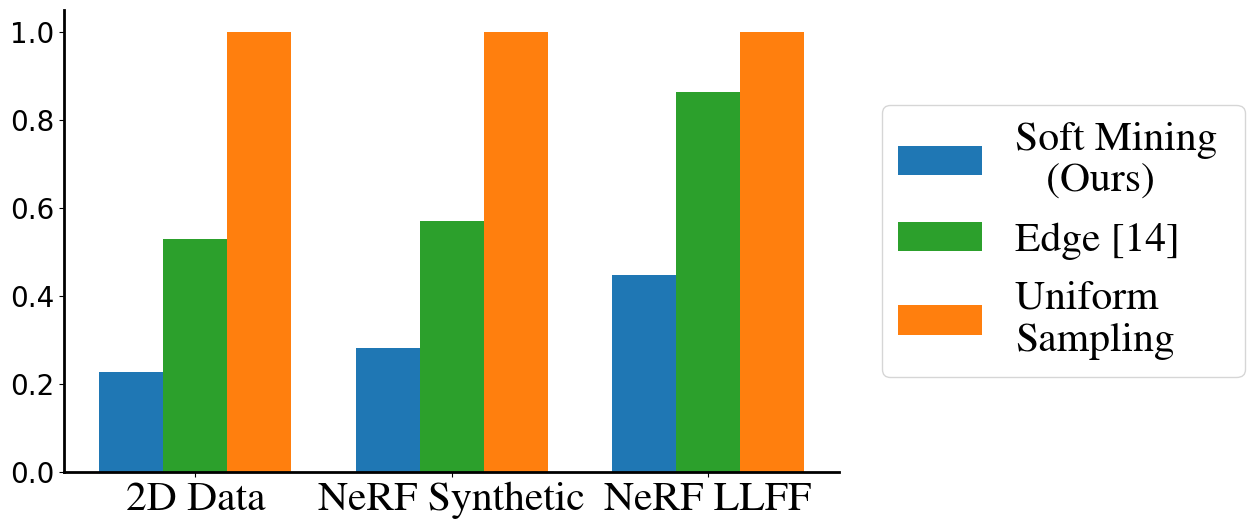}
    \captionsetup{skip=3pt} 
    \caption{
    {\bf Convergence}:
    we show the relative number of iterations compared to uniform sampling required to reach PSNR value of 35 dB for 2D image fitting, 30 dB for the NeRF Synthetic dataset, and 25 dB for the NeRF LLFF dataset.
    Our method requires significantly less number of iterations, specifically less than half of what is required with uniform sampling.
    }
    \label{fig:converge_iter}
\end{figure}

To verify the efficacy of our method, we apply it to two common use cases of neural fields: 2D image fitting and NeRF.
As shown in \cref{fig:converge_iter}, our method at least \textit{doubles} the convergence speed for all tasks.
Compared to the edge-based heuristic~\cite{GAI2023104670} for NeRF, our method approximately doubles the convergence speed for both synthetic and real-world data.

\section{Related Works}
Neural fields have gained substantial popularity as a representation method for various types of signals in a wide variety of applications~\cite{neuralfields}.
This paradigm was first popularized as a way of representing 3D scenes and objects, beginning with Scene Representation Networks~\cite{sitzmann2019srns} in the form of learnable ray marching, which was soon followed by Neural Radiance Fields~\cite{Mildenhall2020nerf} and Implicit Differentiable Renderer~\cite{yariv2020multiview}, which adopted volumetric and distance field representations respectively.
Initially, work on neural scene representations focused on architectures where the field coordinate is a spatial position, but later works generalized this to cases where the parameters of a viewing ray are used to define the field~\cite{sitzmann2021lfns, srt22}.
These methods share a common photometric reconstruction strategy for supervision in which the network is trained to reproduce pixel values of the training images when rendered.
Due to the computational expense of rendering, this supervision is implemented as a stochastic approximation in which mini-batches of pixels are sampled from the training images to be used as supervision at each step.
It is this pixel sampling process that we aim to improve, so to achieve better training convergence and higher accuracy.

\paragraph{Accelerating neural fields.}
Neural scene representations, particularly those based on volume rendering, quickly gained a reputation for being very computationally expensive and therefore slow to train.
Consequently, a number of methods have been proposed for accelerating both training and inference by modifying the model architecture to be less expensive.
The majority of these approaches have tried to achieve their speed-ups by reducing the number of times the underlying field needs to be evaluated \cite{nsvf, yu_and_fridovichkeil2021plenoxels, yu2021plenoctrees, mueller2022instant}, by altering the memory-computation trade-off of the neural architecture \cite{rebain2021derf, reiser2021kilonerf, garbin2021fastnerf, yu_and_fridovichkeil2021plenoxels, sun2022direct}, and/or by optimizing their implementations to take maximum advantage of acceleration hardware with efficient compressed field representations \cite{mueller2022instant, chen2022mobilenerf, tensorrf, eg3d, cao2023hexplane, fridovich2023k}.
While these methods are largely aligned with our goal of improving efficiency, and thereby quality achievable within a given compute budget, their methods are \textit{complementary} to ours, which is compatible with a wide variety of efficient architectures.
We note also that none of these methods focus on choosing which rays~(queries) to use for training, which is the primary focus of our work.

\paragraph{Efficient selection of queries.}
Some prior works have explored modified ray sampling schemes for training NeRF models.
\citet{GAI2023104670}, which we refer to as Edge, samples rays based on detected edges in training images, taking advantage of the fact that rendering error tends to be higher in these regions.
\citet{zhang2022fast} adopts a strategy that incorporates a prior probability derived from local color variance, and further tracks photometric error throughout training with an adaptive quadtree structure, allocating more samples to regions with higher error.
Unlike these hand-crafted techniques, our method seeks to improve training by allocating samples in a principled manner, and in a way that is not tightly coupled to a specific task.
By doing so we show that a significant speed up in convergence is achievable compared to existing methods.
\section{Method}
Let us start by formalizing neural field training~(\cref{sec:formalize}).
We will then introduce our soft mining approach~(\cref{sec:softmining}) as well as how to create batches effectively with minimal overhead via LMC sampling~(\cref{sec:lmc}).

\subsection{Neural field training}
\label{sec:formalize}
A neural field $f_{\params}$ with learnable parameters $\params$ defines a mapping from a bounded set of coordinates $\support \subset \real^D$ to outputs $\mathcal{O} \in \real^F$ as $f_{\params}: \real^D \rightarrow \real^F$, 
where for example $f_{\params}: \real^3 \rightarrow \real^1$ maps positions to signed distance in~\cite{Park_2019_CVPR, sitzmann2020siren}, and $f_{\params}: \real^5 \rightarrow \real^4$ maps positions to view-dependent radiance and density in~\cite{Mildenhall2020nerf}.
These neural fields are typically trained with a loss function that is associated with the error that the neural field function is making while predicting a ground-truth signal $\gt(\coord)$. 
Formally, we define an error function that takes a neural field~$f_{\params}(\coord)$ at the coordinates $\coord$ and outputs a value representing some disparity metric between the prediction and ground truth $\gt(\coord)$ as 
\begin{equation}
\error(\coord) = \error(f_{\params}(\coord), \gt(\coord)) \in \real.
\end{equation}
We can then write the loss function that is minimized for training to be the Monte Carlo estimate of the expectation of this error $\error(\coord)$:
\begin{align}
    \loss 
    &= 
    \frac{1}{N} \sum_{n=1}^{N} \error(\coord_n)
    \approx  
    \expect_{\coord \sim \sampledist(\coord)} \left[\error(\coord)\right] 
    \nonumber
    \\
    &= 
    \int \error(\coord) \sampledist(\coord) \: d\coord
    \;,
    \label{eq:minibatch}
\end{align}
where $N$ is the number of training samples in a batch and~$\sampledist$ is the distribution of the sampled data points~$\coord_n$. 
In \cref{eq:minibatch}, it is important to note that $\sampledist$ affects the outcome of the integral and the expectation, and changes in the sampling scheme effectively result in changes in the loss---naively choosing a batch construction scheme can therefore be harmful and requires care.
Typically, a \textit{uniform} distribution is employed as $\sampledist$ for most neural field applications; \eg, in~\citet{Mildenhall2020nerf}.
Let us now discuss how to perform (soft) hard mining with importance sampling and how it tightly relates to the original objective in \cref{eq:minibatch}.

\subsection{Soft mining with importance sampling}
\label{sec:softmining}

We first start by introducing \textit{importance sampling} for neural field training.
To allow for different strategies to be used for batch construction in \cref{eq:minibatch}, we introduce an importance distribution $\importance(\coord)$  and create training batches as:
\begin{align}
    \int \error(\coord) P(\coord) \: d\coord &= \int \frac{\error(\coord)P(\coord)}{\importance(\coord)} \importance(\coord) \: d\coord
    \nonumber
    \\
    &= \expect_{\coord \sim \importance(\coord)}\left[ \frac{\error(\coord)P(\coord)}{\importance(\coord)} \right].
    \label{eq:impsampwithp}
\end{align}
When $P$ is a uniform distribution, 
which is reasonable to assume given that we typically want the signal to be well-represented everywhere,
the Probability Density Function (PDF) of the uniform distribution is constant, and we can remove $P(\coord)$ from \cref{eq:impsampwithp} and rewrite it as:
\begin{equation}
    \expect_{\coord\sim\importance(\coord)} \left[\frac{\error(\coord)}{\importance(\coord)}\right]
    .
    \label{eq:impsampleExp}
\end{equation}
We can thus further rewrite \cref{eq:minibatch} as:
\begin{equation}
    \loss 
    = 
    \frac{1}{N} \sum_{n=1}^{N} \frac{\error(\coord_n)}{\importance(\coord_n)}, \quad \text{where}\;\; \coord_n \sim \importance(\coord)
    .
    \label{eq:impsample}
\end{equation}
Note that training with \cref{eq:impsample} is impractical for an arbitrary distribution $\importance(\coord)$ as one must differentiate through the sampling process.
Hence, we simply add a stop gradient operator $\stopgrad(\cdot)$ as in~\citet{vqvae}:
\begin{equation}
    \loss \approx \expect_{\coord\sim\stopgrad(\importance(\coord))}\left[\frac{\error(\coord)}{\stopgrad(\importance(\coord))}\right]
    .
    \label{eq:impsamplesg}
\end{equation}
Note that this effectively amounts to simple sample reweighing.
Without affecting the initial training objective, equation~\cref{eq:impsamplesg} now allows us to focus our training efforts by choosing an appropriate $\importance(\coord)$.
Note also that this loss scaling is \textit{not accounted for} in methods that employ heuristics for constructing training batches~\cite{GAI2023104670, zhang2022fast}.

To focus our training on regions with high error we simply define $\importance(\coord)$ to be \textit{proportional} to $\error(\coord)$.
Among various possibilities, for our experiments, we opt for squared error for $\error(\coord)$, which is the typical choice for training neural fields, and set $\importance(\coord)$ to be the L1 norm of the error, so to avoid focusing too much on outliers:
\begin{align}
    \error(\coord) &= \|f_{\params}(\coord) - \gt(\coord)\|_2^2, 
    \\ 
    \importance(\coord) &= \|f_{\params}(\coord) - \gt(\coord)\|_1
    .
\end{align}

\paragraph{Soft mining.}
While importance sampling mathematically allows us to optimize the original objective with potentially more effective samples at hard regions, our experiments in~\cref{sec:softhardmine} show that purely relying on it does not improve performance drastically.
We also show that ignoring the importance weight ${Q(\coord)}^{-1}$, which is equivalent to the commonly-used hard mining~\cite{shrivastava2016training, simo2015discriminative}, is sub-optimal as well.
While more emphasis should be given to the samples that are `hard' to learn, focusing exclusively on the hard samples biases the training too far away from the original training objective.
To address this issue, we propose an alternative that strikes a middle-ground, and write:
\begin{equation}
    \loss = \frac{1}{N} \sum_{n=1}^{N}  \left[ \frac{\error(\coord_n)}{\stopgrad(\importance(\coord_n))^\hardparam} \right],
    \quad
    \text{where } \hardparam \in [0, 1],
    \label{eq:impsampler}
\end{equation}
where $\hardparam$ controls the `softness' of the mining, with $\hardparam{=}0$ corresponding to (pure) hard mining, and $\hardparam{=}1$ corresponding to (pure) importance sampling.
In our experiments we typically utilize $\hardparam \in [0.6, 0.8]$, thus behaving as soft mining.
We ablate our choice in \cref{sec:softhardmine}.

\subsection{Sampling via Langevin Monte Carlo}
\label{sec:lmc}

To implement our method we require that we can sample from an arbitrary distribution $\importance(\coord)$.
This is non-trivial, and we thus rely on Markov Chain Monte Carlo~(MCMC).
Among the family of MCMC methods, we specifically use Langevin Monte Carlo (LMC)~\cite{mcmchandbook2012,Brosse2018} thanks to its simplicity and effectiveness.
Its \textit{deterministic} nature, driven by the gradient of the log posterior distribution, steers the exploration effectively towards regions of higher probability.
Meanwhile, its \textit{stochastic} nature facilitates a comprehensive exploration of the parameter landscape, aiding in evading local minima and promoting convergence to the target distribution.
Formally, to sample from $\importance(\coord)$, denoting a sample location at sampling step $t$ as $\coord_t$ we write:
\begin{equation}\label{eq:lmcupdate}
    \coord_{t+1} = \coord_{t} + a \nabla \log \importance(\coord_t) + b \noise_{t+1},
\end{equation}
where $a{>}0$ is a hyperparameter defining the step size for the gradient-based walks, and $b{>}0$ is a hyperparameter defining the step size for the random walk $\noise_{t+1} {\sim} \calN(0, \mathbf{1})$.
Note that for simplicity in notation, we have folded the two hyperparameters associated with the random walk in \citet[Eq.~2]{Brosse2018} into a single parameter $b$.
Critically, note here that~\cref{eq:lmcupdate} depends only on the local gradient of the sampling distribution $\importance(\coord)$ and random noise $\noise_{t+1}$.
Because the method is local, we can perform sampling with minimal overhead as we already compute the backward pass for training our neural field.

\paragraph{Sample (re-)initialization.}
While LMC eventually converges to the desired distribution, it is well known that MCMC methods require careful (re-)initialization for effective sampling~\cite{gilks1995markov}.
We first initialize the sampling distribution to be uniform over the domain of interest as $\coord_0{\sim}\mathcal{U}(\support)$.
We further re-initialize samples that either move out of $\support$ or have too low error value causing samples to get~`stuck'.
We use uniform sampling as well as edge-based sampling for 2D workloads.

\begin{figure}
    \centering
    \includegraphics[width=\linewidth]{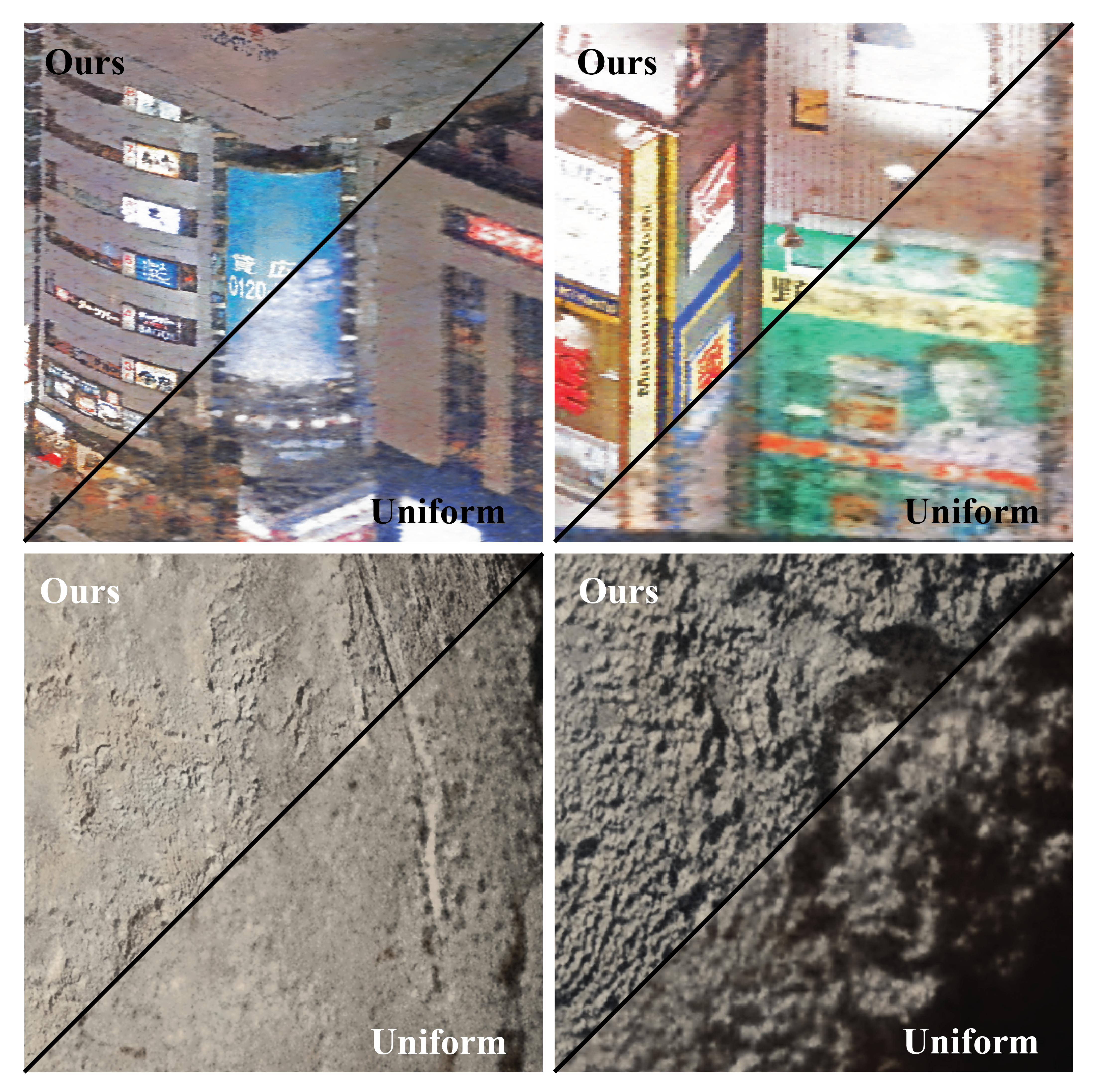}
    \vspace{-2em}
    \caption{
    {\bf 2D image fitting examples}:
    We show 2D image fitting example for our method and uniform sampling for \textbf{(top)} the two regions on the Tokyo image and \textbf{(bottom)} the two regions on the Pluto image.
    The \textbf{(left)} column shows results for the batch size of 256 trained for 10k iterations and the \textbf{(right)} column shows the result for the batch size of 4096 trained for 1k iterations.
    As shown, our results are sharper, especially noticeable around the texts and finer details.
    }
    \vspace{-1em}
    \label{fig:imagefit_qualitative}
\end{figure}

\paragraph{Warming up soft mining.}
We empirically noticed that in very early training iterations ($\le$1k iterations)~LMC requires warm-up time, which is commonly the case for~MCMC methods~\cite{gilks1995markov}.
This makes applying the corrections in \cref{eq:impsampler} unreliable as our samples are not yet exactly following $\importance(\coord)$.
We thus start with $\hardparam{=}0$, \ie, no correction, then linearly increase it to the desired $\hardparam$ value at 1k iterations.

\paragraph{Alternative: multinomial sampling.}
For many neural fields use cases~\cite{sitzmann2020siren, Mildenhall2020nerf}, the modeling space is discrete -- \eg pixels.
In this case, $\importance(\coord)$ would be a multinomial distribution that could be explicitly modeled and sampled from.
While we show in \cref{sec:experiments} that this approach indeed provides an improvement, it is \textit{computationally impractical}.
To use multinomial sampling, one needs to do a forward pass of all data points to build a probability density function, which is computationally expensive.
Even a naive strategy to prevent these forward passes, such as bookkeeping a moving average of error can be costly in a large dataset.
Hence an alternative strategy, such as those based on Markov Chain Monte Carlo (MCMC) is required.

\begin{figure*}
  \centering
    \includegraphics[width=0.32\linewidth]{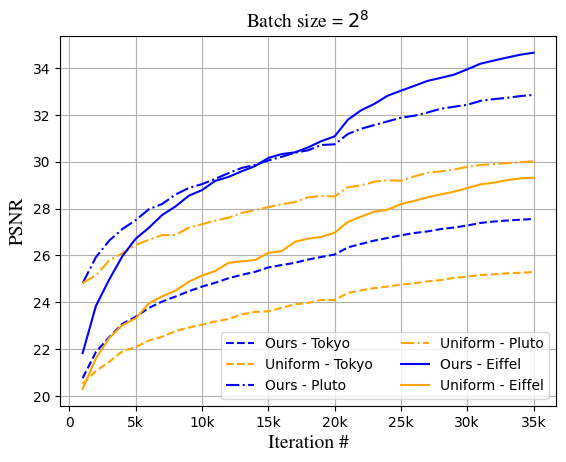}
    \hfill
     \includegraphics[width=0.32\linewidth]{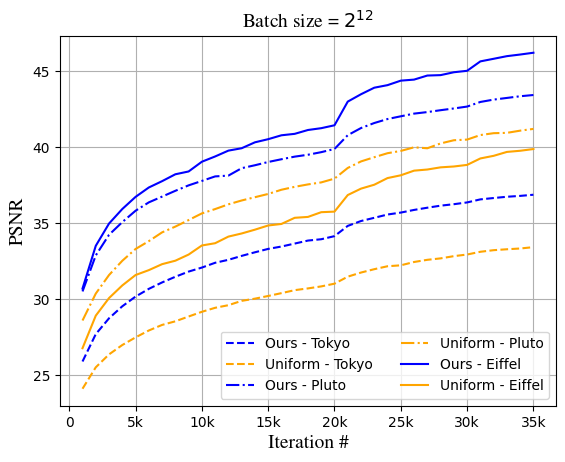}
    \hfill
    \includegraphics[width=0.32\linewidth]{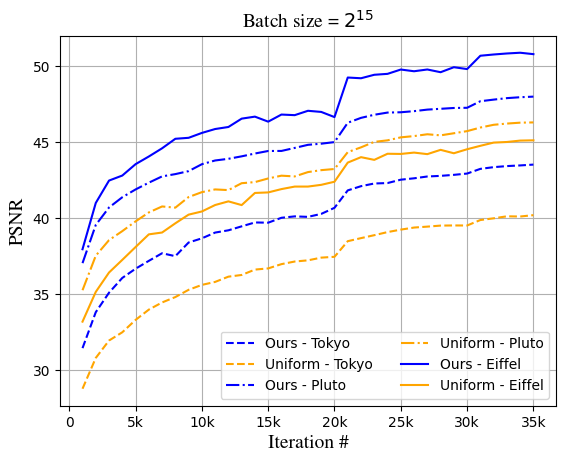}

  \vspace{1em} 

  \setlength{\tabcolsep}{12pt}
  \resizebox{\linewidth}{!}{
    \begin{tabular}{@{}llccccccccccc@{}}
    \toprule
    & batch size     &  $2^8$   & $2^9$   & $2^{10}$   & $2^{11}$   & $2^{12}$   & $2^{13}$   & $2^{14}$   & $2^{15}$   & $2^{16}$   & $2^{17}$   & $2^{18}$   \\
    \midrule
    \multirow{2}{*}{Pluto} & Uniform & 
       30.01 & 32.34 & 35.27 & 38.50 & 41.19 & 43.22 & 45.00 & 46.29 & 47.47 & 48.28 & 48.79 \\
    & Ours & 
       \textbf{32.85}  & \textbf{35.60}  & \textbf{38.55}  & \textbf{41.16}  & \textbf{43.42}  & \textbf{45.33}  & \textbf{46.88}  & \textbf{47.99}  & \textbf{48.88}  & \textbf{49.43}  & \textbf{49.97} \\
    \multirow{2}{*}{Eiffel} & Uniform &  29.32           & 31.96           & 34.59           & 37.21           & 39.87           & 42.11           & 44.07           & 45.12           & 45.45           & 46.00           & 46.14 \\
     & Ours & 
        \textbf{34.65}  & \textbf{37.81}  & \textbf{40.90}  & \textbf{44.00}  & \textbf{46.21}  & \textbf{48.31}  & \textbf{49.62}  & \textbf{50.79}  & \textbf{51.72}  & \textbf{52.33}  & \textbf{52.67} \\
    \multirow{2}{*}{Tokyo} & Uniform & 
    25.30           & 26.95           & 28.87           & 30.95           & 33.41           & 36.07           & 38.56           & 40.19           & 41.46           & 42.15           & 42.66 \\
     & Ours  & 
       \textbf{27.55}  & \textbf{29.51}  & \textbf{31.73}  & \textbf{34.18}  & \textbf{36.86}  & \textbf{39.39}  & \textbf{41.57}  & \textbf{43.52}  & \textbf{45.05}  & \textbf{45.99}  & \textbf{46.62} \\
    \bottomrule
    \end{tabular}
    }

  \caption{ {\bf Convergence -- image fitting:}
  we report the PSNR values for the Pluto, Eiffel Tower, and Tokyo images, with different batch sizes, for both our method and uniform sampling. 
  We also show the convergence graphs for two representative batch sizes.
  Regardless of the batch size, our method provides faster convergence.
  }
  \label{fig:imagefit_convergence}
\end{figure*}

\section{Results}
\label{sec:experiments}

To validate the effectiveness of our method we focus on two popular applications: 2D image fitting and Neural Radiance Fields (NeRF).
We first present results for these two applications and then provide ablation studies.
We note that in all of our experiments, to account for the randomness of neural field training, all reported results are the average outcomes of three runs.
We will release the code to ensure full reproducibility.

\subsection{2D image fitting}
\label{sec:imagefit}
We first apply our method to the task of fitting a neural field to an image, that is, the task of image memorization.
We compare our method to uniform sampling, with the same Instant-NGP~\cite{mueller2022instant} backbone.
We implement our framework based on the official Instant-NGP implementation~\cite{mueller2022instant,tiny-cuda-nn}.

\paragraph{Dataset and metrics.}
We use three high-resolution images for evaluation:
Eiffel Tower~($3024{\times}4032$), Pluto~($8000{\times}8000$) and Tokyo~($6144{\times}2324$).
Pluto image is a high-resolution image of Pluto.\footnote{
Image courtesy of NASA's Photojournal (Image ID: PIA19952).
The image is in the public domain.
}
The latter two were used to benchmark methods in previous works~\cite{mueller2022instant,wu2023neural}.
We compare each method using PSNR.

\paragraph{Hyperparameters.}
We keep the same hyperparameter setting for all our 2D image-fitting experiments.
We use a learning rate of $0.01$ and a multi-step learning rate scheduler (decaying at 20k and 30k iterations) following the base implementation~\cite{mueller2022instant}.
We set $\hardparam{=}0.6$ in \cref{eq:impsampler}.
We leverage image edges for re-initialization~(see \cref{sec:lmc}). 
We execute Sobel edge detection and normalize the edge scores to turn them into a probability distribution.
We then randomly choose pixels from this distribution to re-initialize our LMC samples, those that are the bottom 10\% of the LMC sampling pool according to $\importance(\coord)$.
We further keep 10\% of our training batch to be sampled uniformly to avoid completely ignoring some pixels.
Finally, for \cref{eq:lmcupdate}, we choose $a{=}1e-5$ and $b{=}1e-3$ via hyperparameter search, which we found to work well for most images.

\begin{figure*}
\centering
\includegraphics[width=\linewidth]{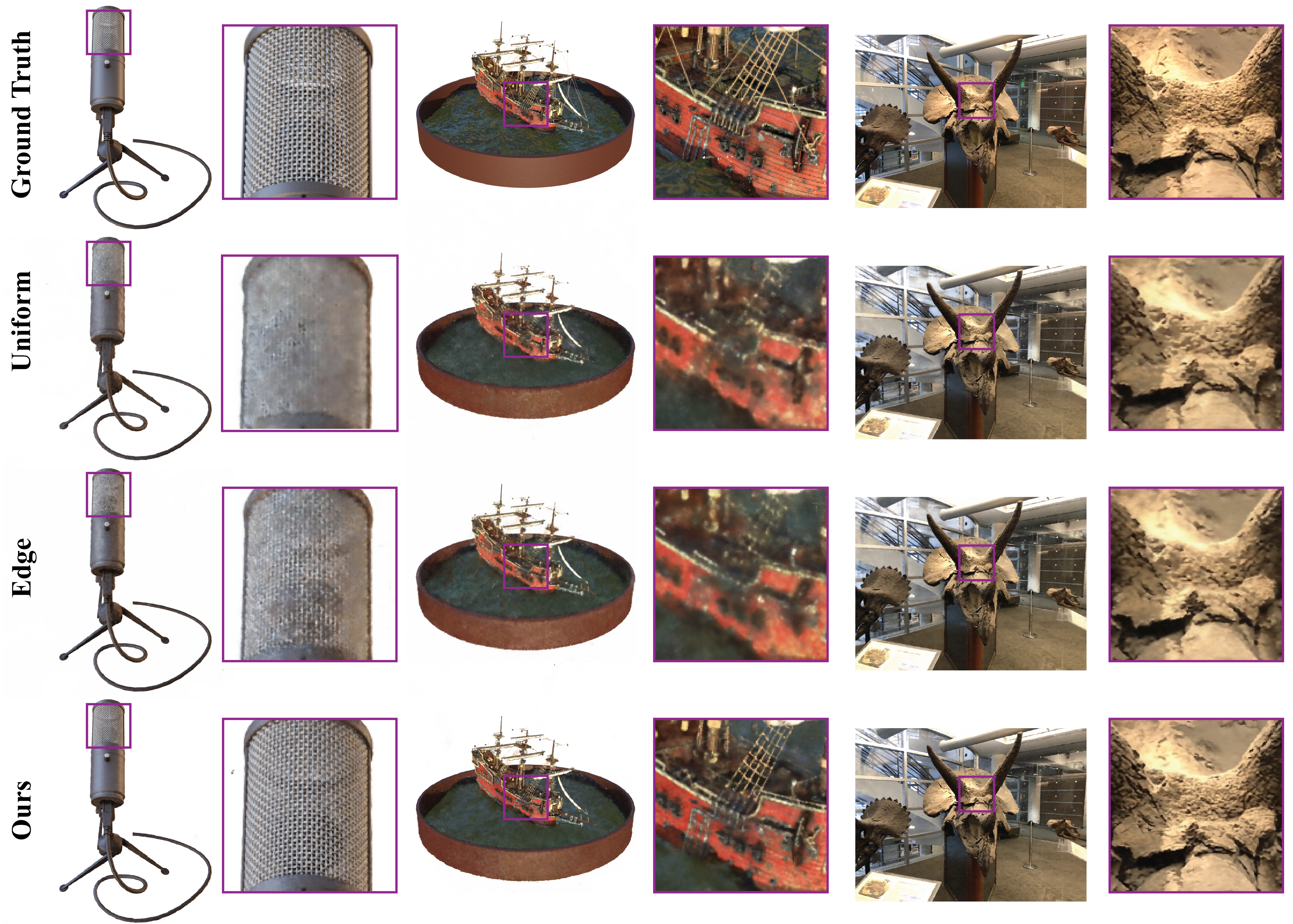}
\vspace{-2em}
\caption{
{\bf Qualitative examples -- NeRF}:
we show example rendering for each sampling method after 1k training iterations.
Our renderings are significantly sharper than other methods, demonstrating faster convergence.
Best viewed when zoomed in.
}%
\vspace{-1em}
\label{fig:nerf_qualitative_all}
\end{figure*}

\paragraph{Results.}
As shown in \cref{fig:imagefit_qualitative}, our method provides significantly higher reconstruction quality for the same number of iterations, \ie, our method converges faster.
We further show convergence graphs in \cref{fig:imagefit_convergence} along with the quantitative results for different batch size configurations.
Notably, our method not only leads to faster convergence but also to higher PSNR at the end. 
We emphasize once more that the only difference between the baselines is the sampling strategy for constructing batches. 
Yet, there is a significant gap, demonstrating the importance of choosing which points to sample.
As depicted in \cref{fig:converge_iter}, we converge almost four times faster than uniform sampling to a~PSNR of 35dB~(averaged over all three images and all batch sizes)---it takes~$\approx$2,400 iterations with our method and~$\approx$10,600 with uniform sampling.

\subsection{Neural radiance fields}

We further experiment on learning NeRF~\cite{Mildenhall2020nerf}, arguably one of the most popular applications for neural fields.
NeRF~\cite{Mildenhall2020nerf} takes a 3D position and a direction vector and outputs radiance and density values used to volume render an image.
NeRF training involves sampling light rays that correspond to each pixel to construct training batches, which are then used to train neural fields with a pixel-wise color reconstruction loss.
We defer the exact details of NeRF to \citet{Mildenhall2020nerf}, as here we are interested only in evaluating how our sampling of the light rays for constructing batches helps convergence.
We compare our method to two baselines: uniform sampling, and Edge~\cite{GAI2023104670}.
As no public implementation exists, we re-implement \citet{GAI2023104670} 
faithfully following the paper.

\paragraph{Dataset and metrics.}
We experiment with the NeRF Synthetic dataset \cite{Mildenhall2020nerf} consisting of 8 object-centric scenes with white background, as well as the LLFF dataset \cite{mildenhall2019llff, Mildenhall2020nerf} consisting of 8 real-world forward-facing scenes.
For the NeRF framework, we use NerfAcc~\cite{li2023nerfacc}, a popular repository which is known to closely reproduce the results of Instant-NGP~\cite{mueller2022instant}~(as the public implementation of \cite{mueller2022instant} does not reproduce the results in the paper for NeRF experiments).
We keep all aspects the same except for the sampling process.
We use the standard image quality metrics: PSNR, SSIM~\cite{wang2004image}, and LPIPS~\cite{zhang2018unreasonable}.

\paragraph{Hyperparameters.}
We train each method with a learning rate of $0.01$, a cosine annealing learning rate scheduler.
We use $\hardparam{=}0.6$ for NeRF Synthetic scenes and $\hardparam{=}0.8$ for the~LLFF dataset, as we found that real and synthetic scenes exhibit different characteristics.
We keep all other hyperparameters the same for all experiments.
We use the same sample re-initialization as in~\cref{sec:imagefit}, and use $a{=}2e1$ and $b=2{e}-2$ for~\cref{eq:lmcupdate}.

\begin{figure*}
    \centering
    \includegraphics[width=0.35\linewidth]{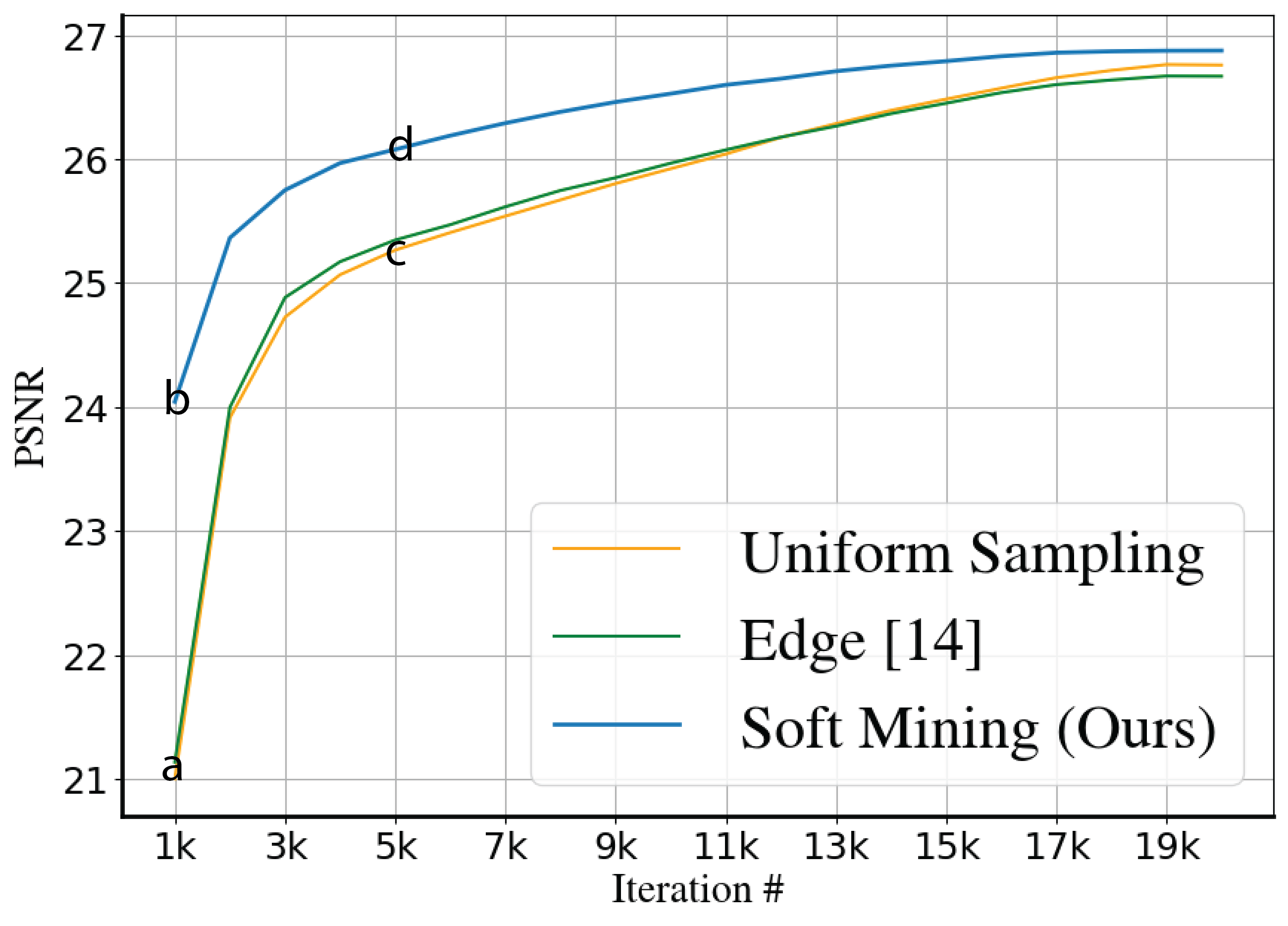}
    \includegraphics[width=0.28\linewidth]{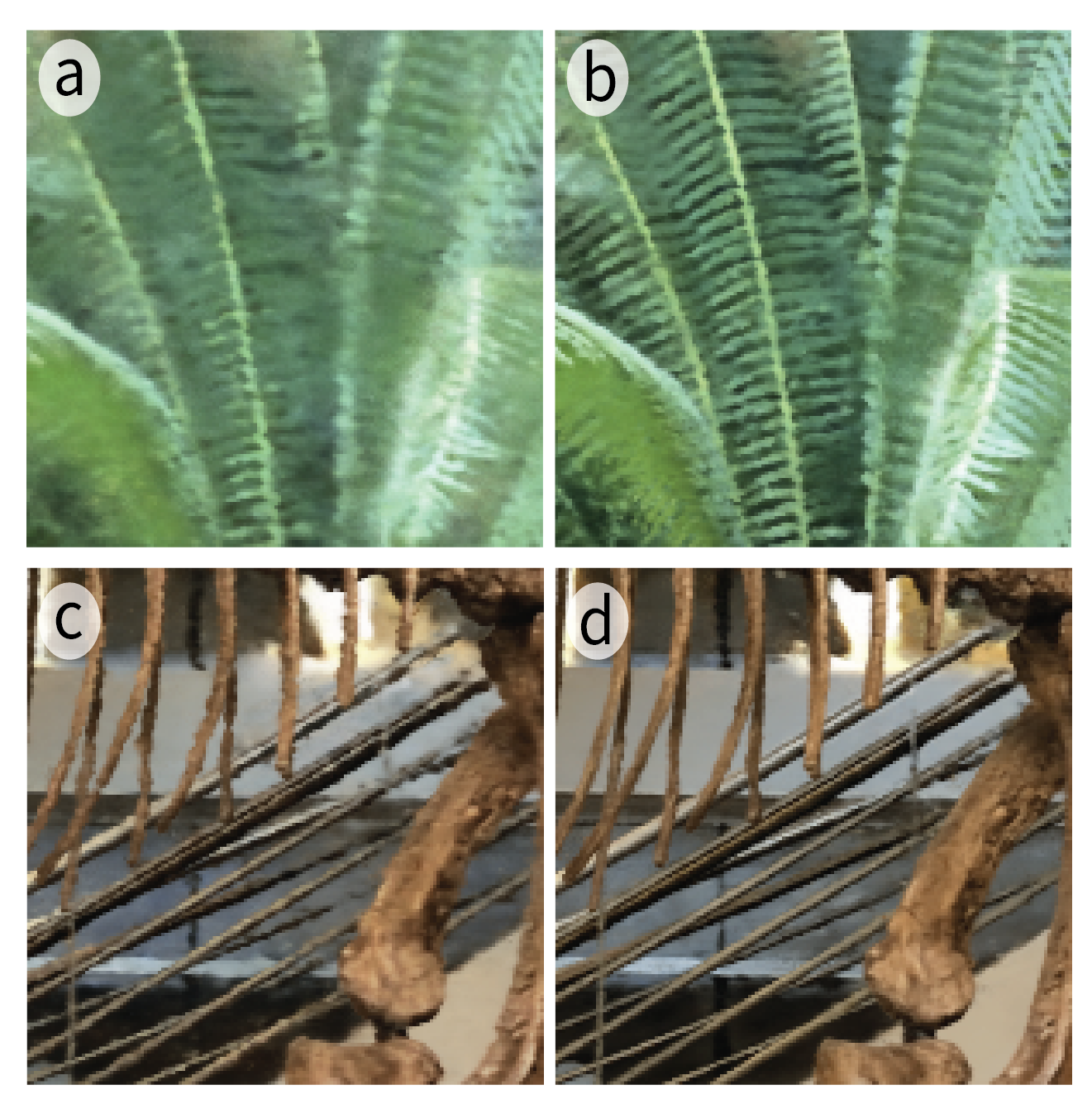}
    \includegraphics[width=0.35\linewidth]{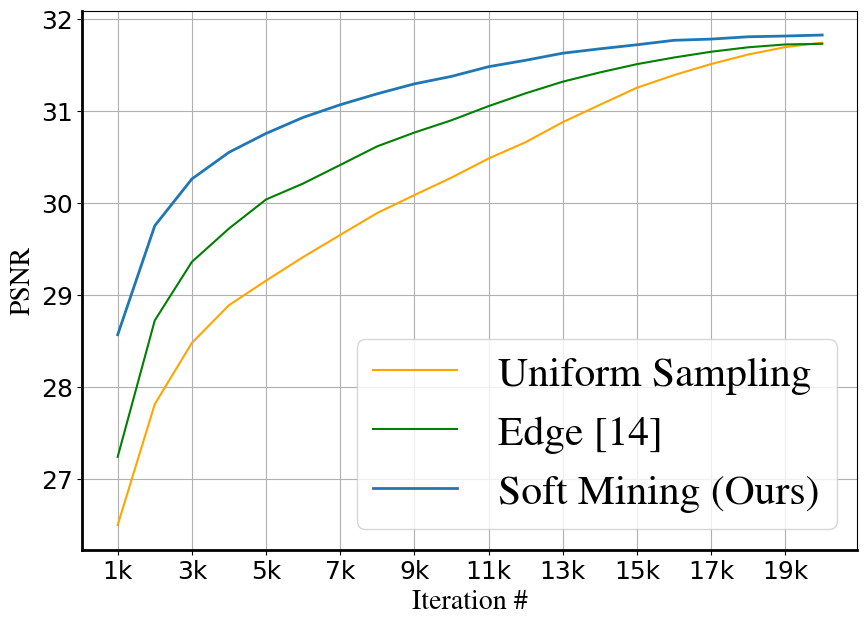}
    \vspace{-1em}
    \caption{
    {\bf Convergence -- NeRF}: 
    PSNR vs. the number of iterations for the test views of {(left)} the LLFF dataset and {(right)} the NeRF Synthetic dataset, with the {(middle)} zoomed-in rendering (fern and trex scenes) for selected iterations.
    Our method significantly speeds up convergence for both datasets, whereas edge-based heuristic~\cite{GAI2023104670} only works well for the synthetic dataset.
    }%
    \label{fig:nerf_convergence_llff}        
    \label{fig:nerf_convergence_synthetic}        

\end{figure*}

\begin{table*}
    \centering
    \resizebox{\linewidth}{!}{
    \setlength{\tabcolsep}{4pt}
    \begin{tabular}{@{}lccccccccccc@{}} \toprule
     & \multicolumn{3}{c}{1k iterations (PSNR / SSIM / LPIPS)} & \scalebox{0.01}{\phantom{abc}}  &  \multicolumn{3}{c}{5k iterations (PSNR / SSIM / LPIPS)} & \scalebox{0.01}{\phantom{abc}}  & \multicolumn{3}{c}{10k iterations (PSNR / SSIM / LPIPS)} \\
    \cmidrule{2-4} \cmidrule{6-8} \cmidrule{10-12}
    & Uniform & Edge \cite{GAI2023104670} & Ours && Uniform & Edge \cite{GAI2023104670} & Ours && Uniform & Edge \cite{GAI2023104670} & Ours \\
    \midrule
    Orchids & 19.03 / 0.56 / 0.44 & 19.14 / 0.56 / 0.45 & \textbf{19.66} / \textbf{0.59} / \textbf{0.41} &  & 20.01 / 0.64 / 0.34 & 20.06 / 0.63 / 0.36 & \textbf{20.14} / \textbf{0.66} / \textbf{0.31} &  & 20.13 / 0.65 / 0.31 & \textbf{20.18} / 0.64 / 0.33 & 20.09 / \textbf{0.67} / \textbf{0.28}  \\
    Trex & 21.38 / 0.75 / 0.41 & 21.52 / 0.76 / 0.41 & \textbf{23.89} / \textbf{0.81} / \textbf{0.37} &  & 25.46 / 0.87 / 0.28 & 26.01 / 0.87 / 0.28 & \textbf{26.67} / \textbf{0.90} / \textbf{0.23} &  & 26.57 / 0.89 / 0.24 & 26.83 / 0.89 / 0.24 & \textbf{27.55} / \textbf{0.91} / \textbf{0.20}  \\
    Leaves & 18.87 / 0.60 / 0.44 & 18.81 / 0.58 / 0.47 & \textbf{19.33} / \textbf{0.61} / \textbf{0.42} &  & 20.51 / 0.68 / 0.33 & 20.40 / 0.66 / 0.37 & \textbf{20.63} / \textbf{0.71} / \textbf{0.30} &  & \textbf{20.85} / 0.71 / 0.30 & 20.73 / 0.69 / 0.34 & 20.82 / \textbf{0.72} / \textbf{0.26}  \\
    Horns & 20.19 / 0.71 / 0.42 & 20.49 / 0.71 / 0.43 & \textbf{23.74} / \textbf{0.75} / \textbf{0.39} &  & 26.21 / 0.82 / 0.29 & 26.18 / 0.81 / 0.30 & \textbf{27.20} / \textbf{0.85} / \textbf{0.24} &  & 27.04 / 0.85 / 0.25 & 26.99 / 0.84 / 0.26 & \textbf{27.94} / \textbf{0.88} / \textbf{0.20}  \\
    Fern & 15.91 / 0.62 / 0.49 & 16.02 / 0.61 / 0.51 & \textbf{23.52} / \textbf{0.72} / \textbf{0.40} &  & 24.49 / 0.78 / 0.31 & 24.35 / 0.76 / 0.34 & \textbf{25.14} / \textbf{0.81} / \textbf{0.29} &  & 25.00 / 0.80 / 0.27 & 24.82 / 0.79 / 0.30 & \textbf{25.42} / \textbf{0.82} / \textbf{0.26}  \\
    Fortress & 21.98 / 0.70 / 0.41 & 23.13 / 0.72 / 0.39 & \textbf{28.49} / \textbf{0.78} / \textbf{0.34} &  & 29.37 / 0.83 / 0.27 & 29.11 / 0.82 / 0.28 & \textbf{30.40} / \textbf{0.86} / \textbf{0.22} &  & 29.92 / 0.85 / 0.23 & 29.78 / 0.84 / 0.24 & \textbf{30.71} / \textbf{0.88} / \textbf{0.19}  \\
    Flower & 24.07 / 0.75 / 0.34 & 24.19 / 0.74 / 0.35 & \textbf{25.49} / \textbf{0.77} / \textbf{0.31} &  & 26.75 / 0.81 / 0.25 & 26.67 / 0.80 / 0.27 & \textbf{27.26} / \textbf{0.83} / \textbf{0.21} &  & 27.39 / 0.83 / 0.22 & 27.42 / 0.83 / 0.23 & \textbf{27.66} / \textbf{0.85} / \textbf{0.18}  \\
    Room & 26.57 / 0.87 / 0.39 & 25.77 / 0.87 / 0.39 & \textbf{28.22} / \textbf{0.90} / \textbf{0.34} &  & 29.33 / 0.92 / 0.28 & 30.01 / 0.93 / 0.26 & \textbf{31.19} / \textbf{0.94} / \textbf{0.21} &  & 30.48 / 0.94 / 0.23 & 30.98 / 0.94 / 0.22 & \textbf{32.05} / \textbf{0.95} / \textbf{0.18}  \\
    \midrule
    Average & 21.00 / 0.70 / 0.42 & 21.13 / 0.69 / 0.43 & \textbf{24.04} / \textbf{0.74} / \textbf{0.37} && 25.27 / 0.79 / 0.29 & 25.35 / 0.79 / 0.31 & \textbf{26.08} / \textbf{0.82} / \textbf{0.25} && 25.92 / 0.81 / 0.26 & 25.97 / 0.81 / 0.27 & \textbf{26.53} / \textbf{0.83} / \textbf{0.22} \\
    \bottomrule
    \toprule
    Mic & 28.29 / 0.95 / 0.08 & 29.55 / 0.96 / 0.08 & \textbf{30.90} / \textbf{0.97} / \textbf{0.06} &  & 31.16 / 0.97 / 0.05 & 32.79 / \textbf{0.98} / 0.04 & \textbf{33.94} / \textbf{0.98} / \textbf{0.03} &  & 32.61 / 0.98 / \textbf{0.03} & 33.96 / 0.98 / \textbf{0.03} & \textbf{34.66} / \textbf{0.99} / \textbf{0.03}  \\
    Ship & 24.78 / 0.80 / 0.28 & 25.28 / 0.80 / 0.30 & \textbf{25.99} / \textbf{0.83} / \textbf{0.26} &  & 27.61 / 0.85 / 0.20 & 27.96 / 0.85 / 0.20 & \textbf{28.33} / \textbf{0.87} / \textbf{0.18} &  & 28.66 / 0.86 / 0.18 & 28.81 / 0.86 / 0.18 & \textbf{29.06} / \textbf{0.87} / \textbf{0.17}  \\
    Lego & 27.04 / 0.90 / 0.14 & 27.81 / 0.91 / 0.15 & \textbf{29.58} / \textbf{0.94} / \textbf{0.09} &  & 30.44 / 0.95 / 0.08 & 31.69 / 0.95 / 0.08 & \textbf{32.70} / \textbf{0.97} / \textbf{0.04} &  & 32.06 / 0.96 / 0.06 & 32.90 / 0.96 / 0.06 & \textbf{33.85} / \textbf{0.97} / \textbf{0.03}  \\
    Chair & 28.62 / 0.93 / 0.10 & 29.50 / 0.94 / 0.10 & \textbf{31.12} / \textbf{0.96} / \textbf{0.07} &  & 31.41 / 0.96 / 0.06 & 32.42 / 0.97 / 0.06 & \textbf{33.46} / \textbf{0.98} / \textbf{0.04} &  & 32.60 / 0.97 / 0.05 & 33.31 / 0.97 / 0.05 & \textbf{34.15} / \textbf{0.98} / \textbf{0.03}  \\
    Materials & 24.21 / 0.88 / 0.16 & 24.34 / 0.87 / 0.19 & \textbf{25.22} / \textbf{0.90} / \textbf{0.14} &  & 26.13 / 0.91 / 0.12 & 26.69 / 0.91 / 0.12 & \textbf{27.29} / \textbf{0.93} / \textbf{0.10} &  & 27.17 / 0.93 / 0.10 & 27.60 / 0.92 / 0.10 & \textbf{27.87} / \textbf{0.94} / \textbf{0.09}  \\
    Hotdog & 31.02 / 0.95 / 0.12 & 31.73 / 0.95 / 0.12 & \textbf{33.44} / \textbf{0.96} / \textbf{0.10} &  & 33.76 / 0.97 / 0.07 & 34.57 / 0.97 / 0.08 & \textbf{35.59} / \textbf{0.98} / \textbf{0.06} &  & 34.85 / 0.97 / 0.06 & 35.41 / 0.97 / 0.06 & \textbf{36.12} / \textbf{0.98} / \textbf{0.05}  \\
    Drums & 22.83 / 0.89 / \textbf{0.15} & 23.01 / 0.88 / 0.20 & \textbf{23.31} / \textbf{0.91} / \textbf{0.15} &  & 23.93 / 0.91 / \textbf{0.12} & 24.08 / 0.91 / 0.13 & \textbf{24.23} / \textbf{0.92} / \textbf{0.12} &  & 24.43 / 0.92 / \textbf{0.10} & \textbf{24.44} / 0.92 / 0.11 & \textbf{24.44} / \textbf{0.93} / 0.12  \\
    Ficus & 25.19 / 0.92 / 0.15 & 26.71 / 0.93 / 0.19 & \textbf{28.97} / \textbf{0.95} / \textbf{0.11} &  & 28.80 / 0.95 / \textbf{0.06} & 30.09 / 0.96 / \textbf{0.06} & \textbf{30.51} / \textbf{0.97} / \textbf{0.06} &  & 29.81 / 0.96 / \textbf{0.05} & 30.76 / \textbf{0.97} / \textbf{0.05} & \textbf{30.85} / \textbf{0.97} / 0.06  \\
    \midrule
    Average & 26.50 / 0.90 / 0.15 & 27.24 / 0.90 / 0.17 & \textbf{28.57} / \textbf{0.93} / \textbf{0.12} && 29.16 / 0.93 / 0.10 & 30.04 / 0.94 / 0.10 & \textbf{30.76} / \textbf{0.95} / \textbf{0.08} && 30.27 / 0.94 / 0.08 & 30.90 / 0.94 / 0.08 & \textbf{31.38} / \textbf{0.95} / \textbf{0.07} \\
    \bottomrule
    \end{tabular}
    }
    \caption{
    {\bf Convergence -- NeRF}:
    for \textbf{(top rows)} LLFF dataset~\cite{mildenhall2019llff, Mildenhall2020nerf} and the \textbf{(bottom rows)} the Synthetic dataset~\cite{Mildenhall2020nerf}.
    Our method provides best results for all cases for early iterations, and almost every case at 10k iterations, when training nearly converges.
    }%
    \vspace{-1em}
    \label{tab:nerf_quantiative}
\end{table*}

\paragraph{Computation time.} 
Before we dive into the results, we first measure the computation time with and without our method with the NeRF application.
We measure the computation time on a system equipped with an Intel Core i7-11700K @ 2.50GHz CPU and NVIDIA GeForce RTX 3090 GPU, with a batch size of 300 rays, with three different runs, each running 20k iterations. 
With our pure PyTorch implementation, running 20k iterations takes 248 seconds, whereas with our method 257 seconds. 
This amounts to a less than $4\%$ increase in processing time.
Given the more-than-twice increase in convergence speed, we argue that the 4\% increase is negligible.
Furthermore, examination with a GPU profiler reveals that the majority of the increase is due to CPU overhead in the backward pass, suggesting that an implementation using a pre-compiled graph such as TorchScript~\cite{PyTorchTorchScript2023} or JAX~\cite{jax2018github} would eliminate this slowdown.
In other words, a more optimized implementation should be able to reduce the computation load even further.
The only computation time that is added is the LMC update rule and the backward step of the last layer that computes the gradients w.r.t the input coordinates.
Both of these should be insignificant compared to the actual neural field training.

\paragraph{Results.}
\cref{fig:nerf_convergence_llff} and \cref{tab:nerf_quantiative} shows the PSNR, SSIM, and LPIPS values for both the NeRF Synthetic and the LLFF datasets. 
As shown, our method is able to speed up convergence in almost all cases, significantly. 
Note especially the gap in performance in the earlier iterations.
As depicted in \cref{fig:converge_iter}, we more than double the convergence speed compared to uniform sampling and approximately double that of Edge~\cite{GAI2023104670}.
More specifically, to reach a PSNR of 25~dB on the LLFF dataset our method takes $\approx$1,700 iterations, uniform sampling takes $\approx$3,800 iterations, and Edge~\cite{GAI2023104670} takes $\approx$3,300 iterations.
On the NeRF Synthetic dataset to reach a PSNR of 30~dB it takes $\approx$2,400 iterations using our method, $\approx$8,500 iterations with uniform sampling, and $\approx$4,800 iterations with Edge~\cite{GAI2023104670}.
Note that in the case of \cite{zhang2022fast}, another method based on image contexts and quadtree subdivision, their relative convergence gain with respect to uniform sampling is $18\%$ and $15\%$ for the NeRF Synthetic and the LLFF dataset with a final gain of ~0.4 PSNR, which we comfortably outperform.

It is worth noting that while the Edge~\cite{GAI2023104670} baseline provides significantly improved results for the synthetic dataset, it does not perform as well on the LLFF scenes.
We suspect that this is because the synthetic dataset is highly particular in that it is object-centric, with a flat white background, while the LLFF scenes are of natural images, thus with a rich background that can have much texture.
We note here that our findings are different from what is reported in Edge~\cite{GAI2023104670}, as they report $\approx$1~dB improvement on average, mostly coming from the `Horns' sequence (21.24~dB with uniform sampling vs 25.45~dB with {Edge~\cite{GAI2023104670}).
However, with our NerfAcc implementation, uniform sampling already provides 27.04~dB at 10k iterations for this scene, and {Edge~\cite{GAI2023104670} provides 26.99~dB.
This difference could be due to implementation details, but we believe the gap in convergence speed between our method and uniform sampling is large enough, even when considering the reported difference.
Further, our method improves convergence regardless of the data type.

\subsection{Ablation studies}
\paragraph{Effect of the soft mining parameter $\hardparam$.}
\label{sec:softhardmine}
To examine the effect of the soft mining parameter $\hardparam$ in \cref{eq:impsampler}, we look at the gain in PSNR values compared to that of uniform sampling for varying $\hardparam$ in \cref{fig:softhardmining}.
As shown, neither complete hard mining nor pure importance sampling is optimal.
Our method provides an effective compromise.
Also note that the gains are more substantial in earlier stages of training as expected, as they both converge to similar solutions, but ours converges much faster.
\quad
It is also important to notice that with $\hardparam{=}0$, which is equivalent to hard mining, results for the LLFF dataset do not improve---rather it degrades.
As 
already discussed theoretically in \cref{sec:softmining}, this hard mining would be a change of the actual objective being minimized, which could cause this performance degradation.

\begin{figure}
  \centering

  \begin{subfigure}[b]{0.495\linewidth}
    \includegraphics[width=\linewidth]{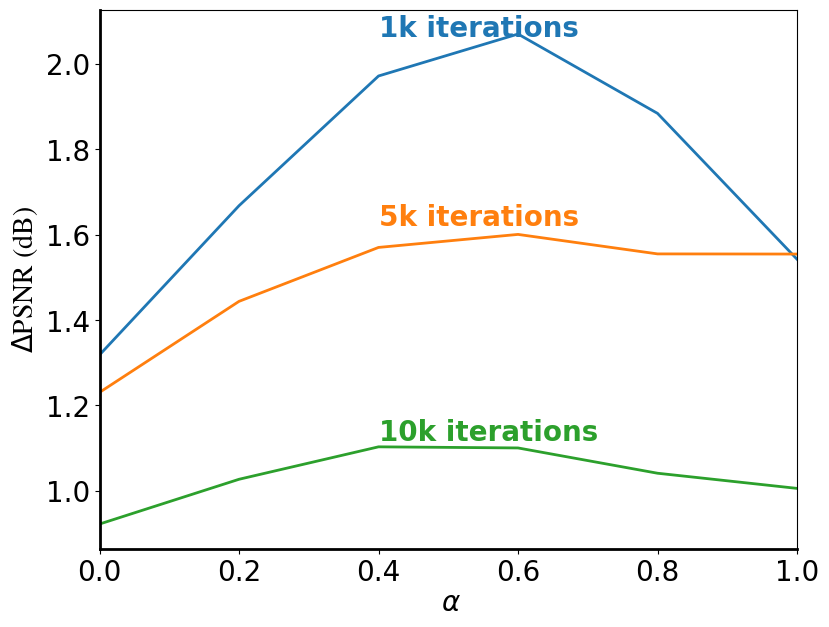}
    \vspace{-1em}
    \caption{NeRF Synthetic Dataset}
    \label{fig:leftimage}
  \end{subfigure}
  \begin{subfigure}[b]{0.495\linewidth}
    \includegraphics[width=\linewidth]{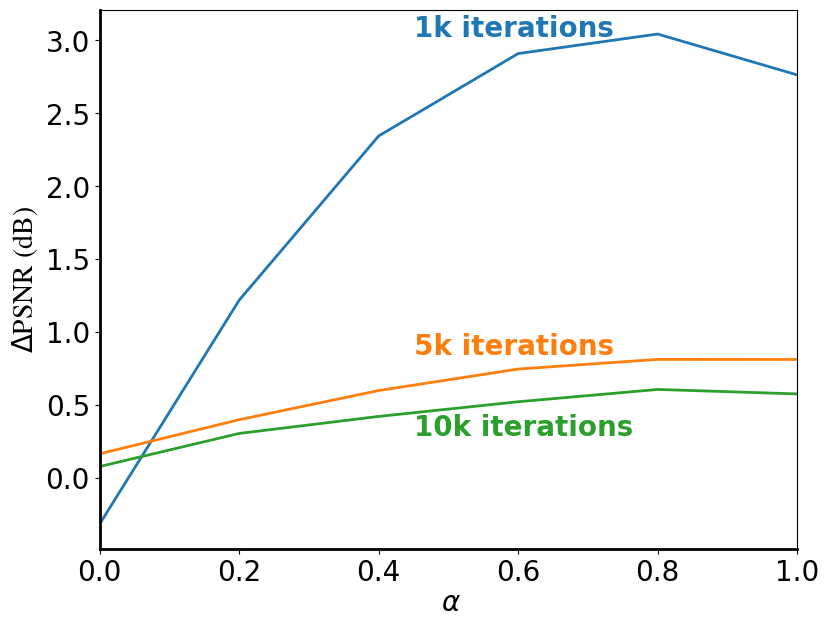}
    \caption{LLFF Dataset}
  \end{subfigure}
  \caption{
  {\bf Ablation -- soft mining parameter $\hardparam$}:
  we report the effect of different $\hardparam$ in terms of PSNR gain compared to uniform sampling for various training iterations.
  In both cases, the optimal choice of $\hardparam$ is within (0,1). 
  Note that simple hard mining, $\hardparam{=}0$ does not improve convergence for the LLFF dataset, whereas with $\hardparam{=}0.8$, our choice, it does.
  }
  \label{fig:softhardmining}
\end{figure}

\paragraph{How effective is Langevin Monte Carlo?}
While we propose to use Langevin Monte Carlo (LMC) to sample with minimal sampling overhead, we also investigate how effective this is compared to the impractical multinomial sampling discussed in \cref{sec:lmc}.
Instead of LMC, at each batch construction step, we evaluate \emph{all pixels} and form the training batch by sampling according to the true importance distribution $\importance(\coord)$, via multinomial sampling.
This is very costly, \eg, increasing the training time to hours or days from minutes depending on the dataset, which destroys any practical gains.
Nonetheless, it can be understood as the upper bound for what can be achieved when infinite compute and resource is available.
Due to heavy computational demand, we only performed this experiment for the Room scene in the LLFF dataset.
We report our results in \cref{fig:ablation_global}.
For both multinomial sampling and LMC, we keep the same hyperparameters as in other experiments, that is,  $\hardparam{=}0.8$.
As shown, by exactly sampling from $\importance(\coord)$ convergence is even higher compared to LMC, as, while samples from LMC theoretically converge to $\importance(\coord)$, with finite samples there is approximation error.
Regardless, even with this error, our LMC sampling performs better than uniform sampling, and provides an effective compromise given the small amount of compute it requires.

\begin{figure}
\centering
\includegraphics[width=0.8\linewidth]{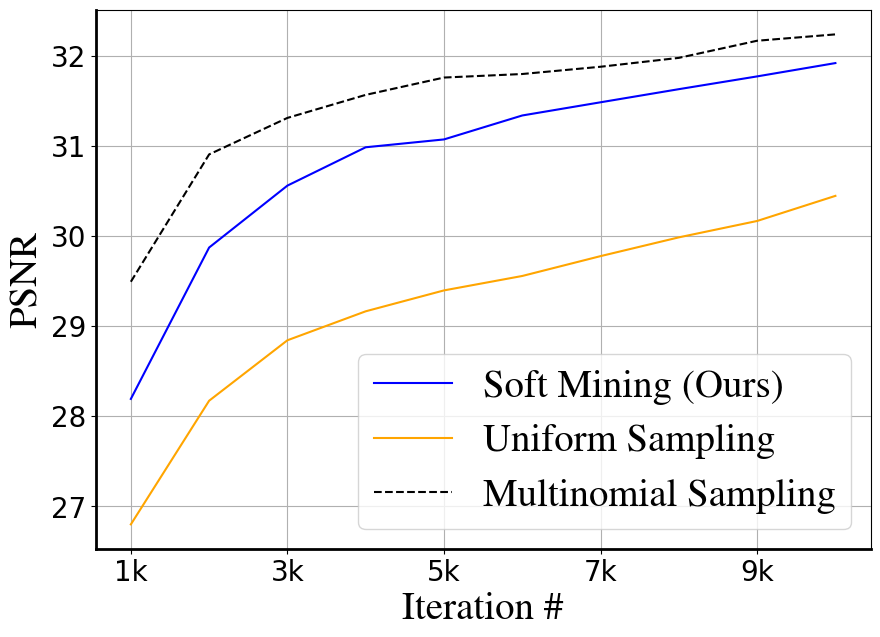}
\caption{
{\bf Soft Mining with LMC vs ideal sampling of $\importance(\coord)$}:
we replace LMC sampling in our method with Multinomial Sampling, which, while computationally heavy, samples directly from $\importance(\coord)$.
Due to the heavy compute of Multinomial Sampling, we only report PSNR curves for the Room scene in the LLFF dataset.
Our method provides an effective compromise with improved convergence with minimal increase in compute.
}
\label{fig:ablation_global}
\end{figure}

\paragraph{How important is (re-)initialization?}
We also validate the importance of uniform samples and re-initialization.
As we report in \cref{tab:ablation_reinit}, both help improve reconstruction quality.
We found them to be particularly useful for achieving good final converged performance.

\begin{table}
\begin{center}
\resizebox{\linewidth}{!}{
\setlength{\tabcolsep}{10pt}    
\begin{tabular}{ccc}
\toprule
w/o Uniform & w/o Re-initialization & with both (Ours) \\
\midrule
23.91 & 23.57 & 24.04 \\
\bottomrule
\end{tabular}
}
\caption{
{\bf Uniform sampling and re-initialization}:
We report PSNR of our method on LLFF dataset after 1k iterations of training, with either uniform sampling or re-initialization disabled.
}
\label{tab:ablation_reinit}
\end{center}
\end{table}

\section{Conclusions}
We presented how to accelerate neural field training by introducing soft mining in the construction of training batches, which we implement via Langevin Monte Carlo.
We have demonstrated on 2D image fitting and NeRF, that our method leads to a two-fold+ improvement in convergence speed.

\paragraph{Limitations and future work.}
While our methods significantly improve results, we still rely purely on the loss function, which may not directly correlate with the application at hand.
In NeRF, for example, training losses may not directly correlate with the novel-view rendering quality. 
As our framework does not depend on the choice of $\importance(\coord)$, it could be possible to perhaps choose a different distribution to sample from, \eg, depending on ray uncertainties~\cite{goli2023bayes} or based on active learning~\cite{choi2021vabal}.
Our method sets a framework that allows easy exploration of such design choices, which was not possible before.

\section{Acknowledgments}
The authors would like to thank Ivan Krasin and David Fleet for their constructive feedback and support of this work. This work was supported in part by the Natural Sciences and Engineering Research Council of Canada (NSERC) Discovery Grant, NSERC Collaborative Research and Development Grant, Google, Digital Research Alliance of Canada, and Advanced Research Computing at the University of British Columbia.

{
    \small
    \bibliographystyle{ieeenat_fullname}
    \bibliography{main}
}

\end{document}